\documentclass[10pt,twocolumn,letterpaper]{article}

\usepackage{iccv}
\usepackage{times}
\usepackage{epsfig}
\usepackage{graphicx}
\usepackage{amsmath}
\usepackage{amssymb}
\usepackage{booktabs}
\usepackage{bm}
\usepackage{multirow}

\usepackage[pagebackref=true,breaklinks=true,letterpaper=true,colorlinks,bookmarks=false]{hyperref}

\usepackage[capitalize]{cleveref}
\crefname{section}{Sec.}{Secs.}
\Crefname{section}{Section}{Sections}
\Crefname{table}{Table}{Tables}
\crefname{table}{Tab.}{Tabs.}

\iccvfinalcopy 

\def\iccvPaperID{6769} 

\ificcvfinal\fi

\begin{document}

\setlength{\abovedisplayskip}{2pt}
\setlength{\belowdisplayskip}{2pt}
\title{Learning Trajectory-Word Alignments for Video-Language Tasks}
\author{Xu Yang$^{1}$ \quad Zhangzikang Li$^{1,2}$ \quad Haiyang Xu$^2$\footnotemark[1] \quad Hanwang Zhang$^3$ \quad Qinghao Ye$^2$ \\ \quad Chenliang Li$^2$
\quad Ming Yan$^2$ \quad Yu Zhang$^1$\footnotemark[1] \quad Fei Huang$^2$ \quad Songfang Huang$^2$ \quad 
\vspace{3pt}\\
\normalsize$^1$Southeast University \quad 
\normalsize$^2$Alibaba Group \quad
\normalsize$^3$Nanyang Technological University\\
\tt\small \{xuyang\_palm, zhang\_yu\}@seu.edu.cn, lizhangzikang@gmail.com, \{shuofeng.xhy, yeqinghao.yqh,\\
\tt\small lcl193798, ym119608, f.huang, songfang.hsf\}@alibaba-inc.com, hanwangzhang@ntu.edu.sg
}
\maketitle
\renewcommand{\thefootnote}{\fnsymbol{footnote}}
\footnotetext[1]{Corresponding authors.}

\begin{abstract}
In a video, an object usually appears as the \textbf{trajectory}, \ie, it spans over a few spatial but longer temporal patches, that contains abundant spatiotemporal contexts. However, modern Video-Language BERTs (VDL-BERTs) neglect this trajectory characteristic that they usually follow image-language BERTs (IL-BERTs) to deploy the patch-to-word (P2W) attention that may over-exploit trivial spatial contexts and neglect significant temporal contexts. To amend this, we propose a novel \textbf{TW-BERT} to learn \textbf{T}rajectory-\textbf{W}ord alignment by a newly designed trajectory-to-word (T2W) attention for solving video-language tasks. Moreover, previous VDL-BERTs usually uniformly sample a few frames into the model while different trajectories have diverse graininess, \ie, some trajectories span longer frames and some span shorter, and using a few frames will lose certain useful temporal contexts. However, simply sampling more frames will also make pre-training infeasible due to the largely increased training burdens. To alleviate the problem, during the fine-tuning stage, we insert a novel Hierarchical Frame-Selector (HFS) module into the video encoder. HFS gradually selects the suitable frames conditioned on the text context for the later cross-modal encoder to learn better trajectory-word alignments. By the proposed T2W attention and HFS, our TW-BERT achieves SOTA performances on text-to-video retrieval tasks, and comparable performances on video question-answering tasks with some VDL-BERTs trained on much more data. The code will be available in the supplementary material.

\end{abstract}

\section{Introduction}
\label{sec:intro}

\begin{figure}[t]
  \centering
   \includegraphics[width=1\linewidth]{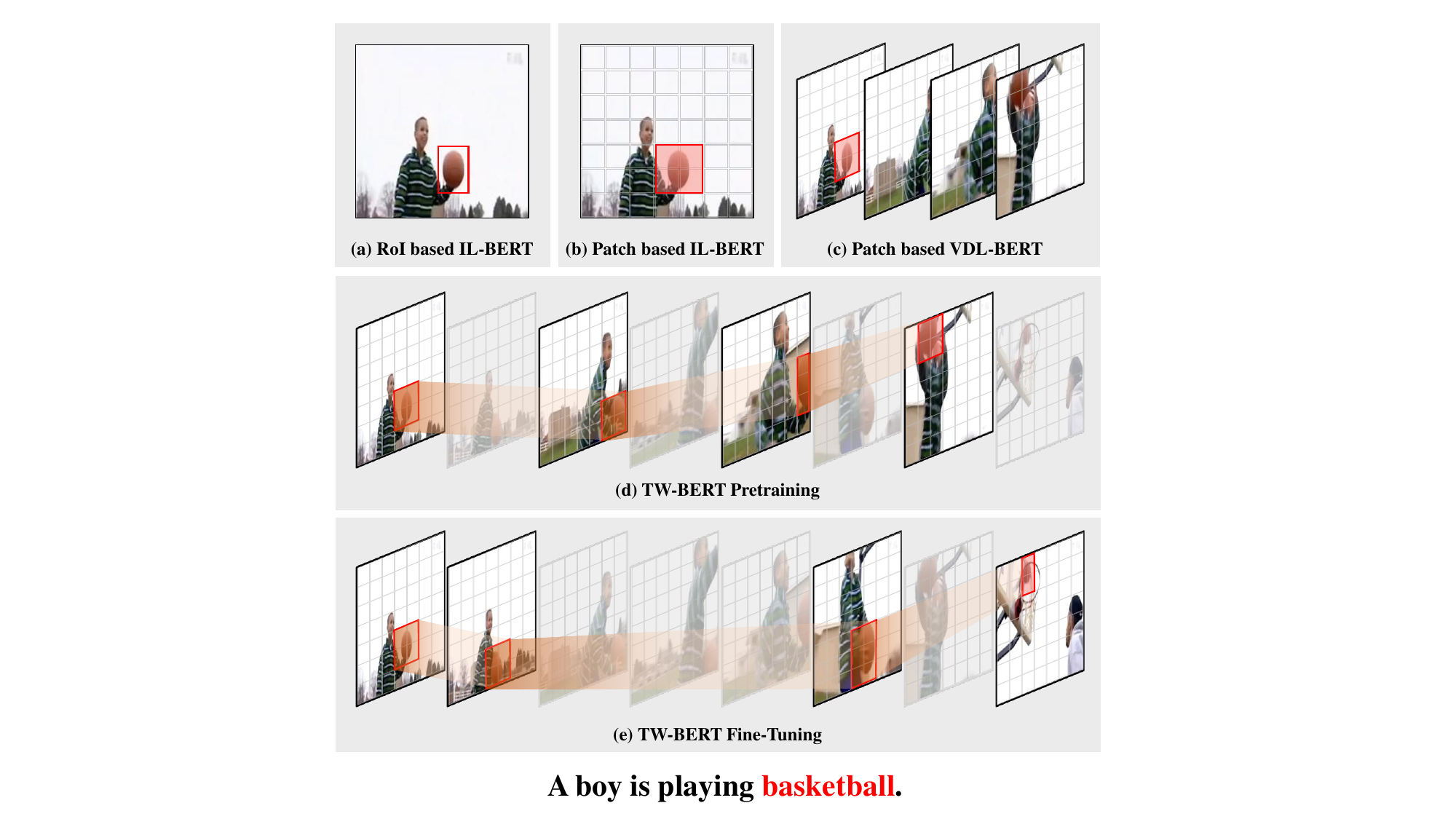}
   \caption{The comparisons of five different ways to build object-word alignments. In (b), by patch-to-word (P2W) attention, implicit object-word alignment can be built, \eg, the ``ball region'' is aligned with \textcolor{red}{basketball}. While in (c), P2W attention may concentrate the attention of an object on only one frame, \eg, \textcolor{red}{basketball} only attends over on the ``ball region'' of the first frame. (d) When pre-training TW-BERT, the trajectory of an object is constructed, \eg, the ``ball trajectory'' through 4 frames built. Note that we only use 4 frames during pre-training, we show 8 frames for comparing with (e) and the transparent ones denote that they are not input into the model. (e) When fine-tuning TW-BERT, we use HFS to select 4 frames from 8 and we can see that the selected frames are different from the uniformly sampled ones in (d).}
   \label{fig:onecol}
   \vspace{-0.2in}
\end{figure}

By witnessing the boom of BERT-like models in single domains, \eg, vision or language\cite{treetran, DistilBERT, ALBERT}, researchers begin to build vision-language BERTs\cite{BEIT, ibot, peco} for learning robust cross-modal associations. Compared with the images, videos provide more details for building robust associations, \eg, the spatiotemporal contexts in videos can better describe the actions. However, directly addressing dynamic videos may cost more storage and computation resources. To circumvent such huge cost, researchers first study image-language BERT (IL-BERT)\cite{ScalingUV, Align, uniter, Oscar, ibot} and then exploit the research fruits to build the challenging, while more pragmatic, video-language BERT (VDL-BERT)\cite{VideoBERT, Frozen, HERO, CLIP4Clip, MultimodalTF, univil, LearningVR, EndtoEndLO, VLM}.

Pioneering techniques are directly inherited from IL-BERT into VDL-BERT, \eg, the image encoders are used to embed sampled video frames~\cite{clipbert,ActBERT}. However, since image encoders can hardly learn temporal contexts, the resultant VDL-BERTs will degrade to IL-BERT even if they are trained by video-text pairs. To ameliorate it, researchers apply the video encoders~\cite{swin, timesformer, VIOLET, ALPRO} to embed spatiotemporal contexts. Though promising improvements are observed, these VDL-BERTs still neglect or ill-consider a significant factor in IL-BERT: the object-word alignment, which helps learn robust associations.

To learn object-word alignments, as Figure~\ref{fig:onecol}(a) shows, researchers use an RoI-based extractor to embed an image into a series of RoI features~\cite{ViLBERT, Oscar, uniter}. However, this RoI-based extractor is offline trained by object detection with limited label inventory, which will weaken the IL-BERT since the extractor will not be updated during large-scale pre-training. To enable the end-to-end training, researchers substitute the RoI-based extractor with visual transformers whose outputs are a series of grid embeddings, which will be used to build the cross-modal connections. Although a single grid usually does not construct an integral object, fortunately, the widely applied patch-to-word (P2W) attention of IL-BERT can softly seek the salient visual regions for a given query word. Then the object-word alignments can still be built between this softly detected object and the word, \eg, as shown in Figure~\ref{fig:onecol}(b), the object ``basketball'' can be implicitly attended by the corresponding word query.

Although the P2W attention remedies the loss of RoI-level features for learning object-word alignments in IL-BERT, its effectiveness is weakened in the video case. This is because the objects usually act as the \textbf{Trajectories} which span a few spatial while multiple temporal grids in the videos. Thus, directly applying the P2W attention may over-exploit the trivial spatial contexts while neglecting the significant temporal contexts and then make the model attend to only one or two frames. Figure~\ref{fig:onecol}(c) shows this limitation that P2W attention only aligns the ``ball'' in the first frame to the word ball.

To address this limitation, we propose to learn \textbf{T}rajectory-to-\textbf{W}ord alignments to solve video-language tasks and name this model as \textbf{TW-BERT}. Specifically, such alignment is learnt by a novel designed trajectory-to-word \textbf{T2W} attention, which first uses the word as the query to seek the salient parts of each frame and the sought parts are sequenced to form the trajectory. Then the query word attends over the trajectories again for capturing cross-modal associations. In this way, the trivial spatial regions are weakened and the temporal contexts will be strengthened, \eg, as shown in Figure~\ref{fig:onecol}(d), the attention weights of the word will be concentrated on the object trajectory instead of only one frame as in (c).  In the implementation, we follow most VDL-BERTs to set up the network: two single-modal encoders for the video and text and one cross-modal encoder, which is sketched in Figure~\ref{fig:framework}. For the cross-modal encoder, since our T2W attention does not have the same structure as the word-to-patch (W2P) attention, our cross-modal encoder is asymmetric. 

Moreover, previous VDL-BERTs usually uniformly sample a few frames into the model, which contains two drawbacks. Firstly, using a few frames may lose temporal context and secondly, uniform sampling can hardly capture the varying graininess of the trajectories, \ie, some trajectories span longer frames and some span shorter. However, simply sampling more frames will largely increase the pre-training burdens that are beyond the computation resources we own. To alleviate this problem, \textbf{in the fine-tuning stage}, we sample more frames into the video encoder while only keeping the most relevant frames according to the corresponding text by a novel designed \textbf{H}ierarchical \textbf{F}rame-\textbf{S}elector (\textbf{HFS}). For example, as shown in Figure~\ref{fig:onecol}(e), 4 frames are selected by HFS from the 8 uniformly sampled ones for learning trajectory-word alignment. Specifically, HFS inserts a few lightweight layers into the video encoder and these layers can gradually filter frames conditioned on the language context. In this way, HFS learns the coarse-grained trajectory-word alignments, \ie, frame-word alignments, to help the later T2W attention learn more fine-grained trajectory-word alignments.


To sum up, our contributions are:

\begin{itemize}
	\setlength{\itemsep}{3pt}
	\setlength{\parsep}{3pt}
	\setlength{\parskip}{3pt}

	\item 
	\vspace{-6pt}
	We propose a novel perspective to consider the videos that are composed of moving object trajectories, which may inspire the researchers to build more advanced VDL-BERTs.
	\vspace{-6pt}
	\item We propose a simple while effective \textbf{T2W} attention to learn \textbf{T}rajectory-to-\textbf{W}ord alignments. 
	\vspace{-6pt}
	\item We propose a novel hierarchical frame-selector (\textbf{HFS}) in TW-BERT during fine-tuning to capture the varying graininess of the trajectory while not largely increasing the training burdens. 
	\vspace{-6pt}
    \item We achieve SOTA performances compared with other VDL-BERTs trained by the same amount of data.

\end{itemize}

\section{Related Work}
\label{sec:related work}
\subsection{Image-Language BERT (IL-BERT)}
Recently, various techniques have been proposed in IL-BERT to learn vision-language connections. Most of them aim at capturing robust object-word alignments since such alignments construct the foundations of visual reasoning. In the beginning, a pre-trained Faster-RCNN~\cite{Faster-RCNN} is used to extract a series of RoI poolings, where each one contains one or a few salient objects, to facilitate building object-word alignments~\cite{ViLBERT, lxmert, uniter, Oscar}. However, this Faster-RCNN is usually pre-trained by the object annotations from COCO~\cite{COCO} and VG~\cite{VG}, whose concept space is much narrower than the data used to train IL-BERT, which can be almost unlimitedly collected from the websites. Moreover, this Faster-RCNN is not updated during the end-to-end training of the IL-BERT, which means that the visual encoder may hardly learn novel knowledge from the web-collected data. Thus the performances of these IL-BERTs are limited.

To further release the potential of hugely web-collected data, the offline Faster RCNN is switched into vision Transformers~\cite{vit, swin} and thus a homogeneous IL-BERT, \ie, all the components are transformer-based, is built, which is more easily trained end-to-end~\cite{ViLBERT}. Compared with the RoI-based encoder, the vision Transformer outputs a series of patch embeddings and thus may lose object-level knowledge. To remedy such loss, various strategies are proposed to improve the vision-language alignments. For example, the align-before-fuse strategy~\cite{Align} aligns the paired image-text embeddings before cross-modal fusion for facilitating the subsequent fusion. And the fine-grained contrastive objective~\cite{FILIP} amplifies the local details for learning the object-word alignments. 

\subsection{Video-Language BERT (VDL-BERT)}
Since the spatiotemporal contexts in videos can hardly be learnt by image extractors, only inheriting the techniques which are successful in IL-BERT to VDL-BERT is not enough. Thus, based on the fruits of IL-BERT, most VDL-BERTs aim at exploiting more spatiotemporal contexts to build cross-modal associations. One straightforward way is to learn such spatiotemporal contexts through video Transformers~\cite{timesformer, mvit}. Besides this, some more advanced techniques are proposed, \eg, VIOLET~\cite{VIOLET} tokenizes the dynamic video patches and predicts the labels of these tokens; BridgeFormer~\cite{BridgeFormer} erases the words (nouns or verbs) from the text and learn to match the visual embeddings queried by the erased words and the remained texts; or ALPRO~\cite{ALPRO} computes the similarities between the video embeddings with the generated entity prompts. TS2-Net~\cite{TS2} shifts token features and selects informative tokens in both temporal and spatial dimensions to produce the fine-grained spatial-temporal video representation. Although substantial improvements are observed, these methods use video patches in the cross-modal encoder, which neglects that an object usually acts as the trajectory in the videos, and thus they may over-exploit the trivial spatial contexts. To ameliorate this limitation, we propose TW-BERT to capture trajectory-word alignments for more robust vision-language alignment.
\begin{figure}[t]
  \centering
   \includegraphics[width=1\linewidth]{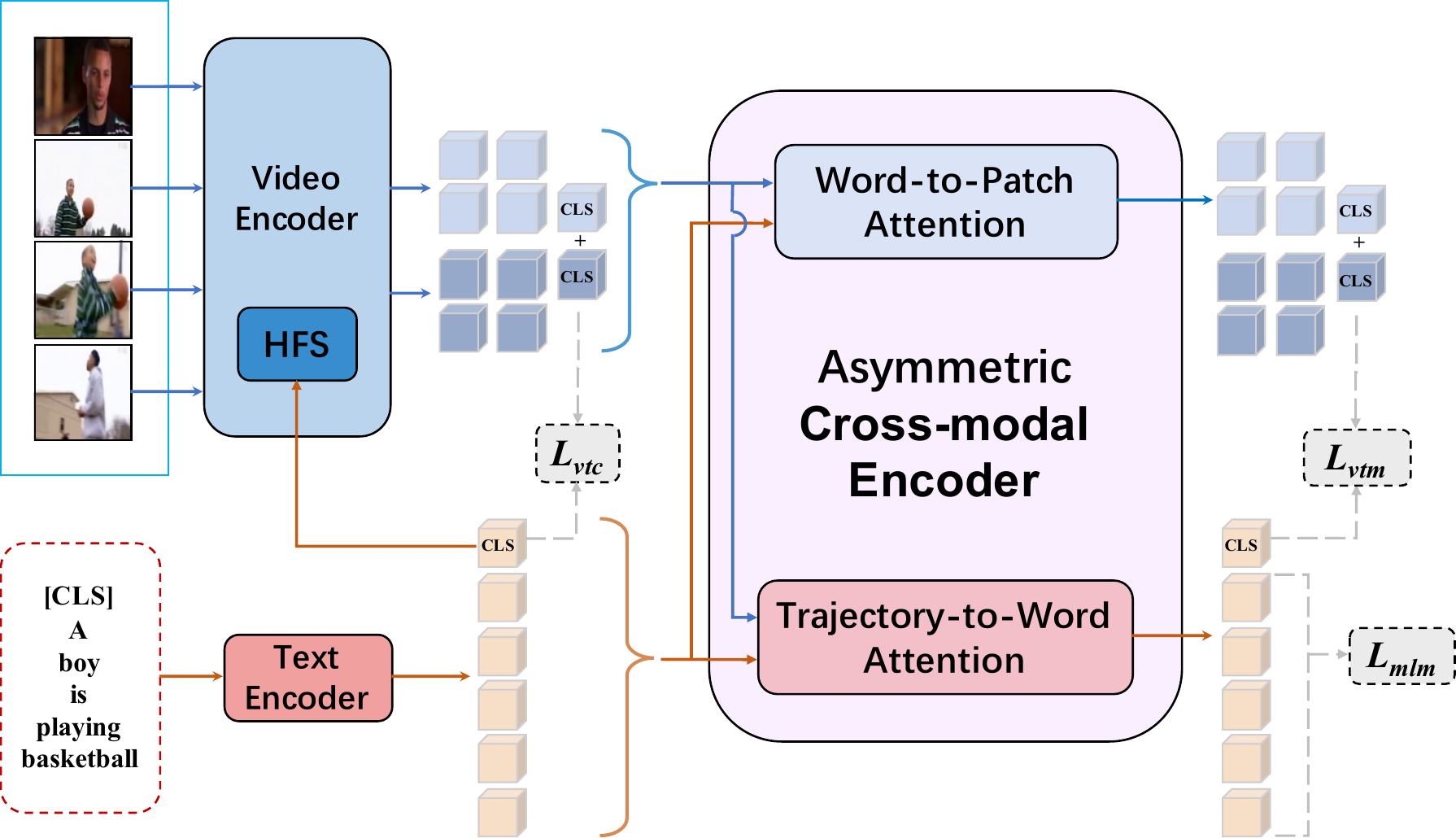}
   \caption{The architecture of TW-BET, which contains two single-modal encoders and one asymmetric cross-modal encoder. Totally three losses in the grey blocks are used to train the whole model: $L_{vtc}$, $L_{mlm}$, and $L_{vtm}$. Note that HFS is only used in the fine-tuning stage.}
   \label{fig:framework}
   \vspace{-0.2in}
\end{figure}

\section{TW-BERT}
\label{sec:approach}
Figure~\ref{fig:framework} sketches TW-BERT, which has two single-modal encoders for embedding the video and text  (cf. Sec.~\ref{subsec:single-modal-encoders}) and one cross-modal encoder for learning video-language associations  (cf. Sec.~\ref{subsec:cross-modal-encoders}). Different from the previous VDL-BERTs, our cross-modal encoder is an asymmetric one that contains a classic word-to-patch (W2P) attention and a novel proposed trajectory-to-word (T2W) attention (cf. Sec.~\ref{subsec:cross-modal-encoders}) for learning trajectory-word alignments. In Sec.~\ref{subsec:hierarchical_frame_selector}, we introduce how to use Hierarchical Frame-Selector (HFS) in the fine-tuning stage to gradually filter the frames for capturing coarse-grained frame-word alignments. Lastly, we introduce the losses used to train our TW-BERT in Sec.~\ref{sec:training_objectives}.

\subsection{Single-Modal Encoders}
\label{subsec:single-modal-encoders}
\noindent\textbf{Video Encoder.}
For a video, we sample a few $224 \times 224$ frames and input them into the 12-layer TimeSformer~\cite{timesformer, ALPRO} for embedding. TimeSformer first partitions each frame into $14 \times 14$ non-overlapping patches, which are flattened and fed to a linear projection layer to produce a sequence of patch tokens. Then TimeSformer applies self-attention along the temporal and spatial dimensions to calculate per-frame features. These features are further mean-pooled along the height and width dimensions. Learnable spatial positional embeddings are added to each video token in the same spatial location of different frames. The final output embedding set is $\bm{V}=\{\bm{v}_{cls},\bm{v}_{1},...,\bm{v}_{N_V-1}\}$, where $\bm{v}_n \in \mathbb{R}^d$ and $\bm{v}_{cls}$ is the global [CLS] embedding.

\noindent\textbf{Text Encoder.}
For a text, we use a 6-layer transformer to embed it and the output is $\bm{X}=\{\bm{x}_{cls},\bm{x}_1,...,\bm{x}_{N_X-1}\}$, where $\bm{x}_n \in \mathbb{R}^d$ and $\bm{x}_{cls}$ is the global [CLS] embedding. Similar to the video encoder, we also add positional embeddings to the text tokens.

\subsection{Asymmetric Cross-Modal Encoder}
\label{subsec:cross-modal-encoders}
After embedding videos and texts, a cross-modal encoder is used to fuse them by calculating bi-directional associations: vision-to-language and language-to-vision. No matter what the direction is, the motivation is to assign semantic knowledge from one domain to another. Since a single word contains integral semantic knowledge, we follow previous VDL-BERT~\cite{ALPRO, Frozen, clipbert, VideoBERT} to set a traditional word-to-patch (W2P) attention to assign the words to a patch. However, different from the word, only the object instead of one grid conveys integral semantic knowledge. In videos, an object usually spans both spatial and temporal axes and thus the previously used patch-to-word (P2W) attention may fail to transfer the semantic knowledge. To amend this, we design a novel Trajectory-to-Word (T2W) attention for transferring semantic knowledge from videos to texts. Since W2P and T2W attentions have diverse structures, our cross-encoder is asymmetric.

Specifically, both W2P and T2W attentions are built on the Multi-Head Attention (MHA) operation, here we first formalize MHA and then introduce how to use it to build W2P and T2W attentions. Formally, MHA is\footnote{To avoid symbol confusions, we use the calligraphic font ($\setminus$mathcal commend in LaTex) to denote the built-in variables of the MHA module.}:
\begin{equation} \label{equ:multi-head}
\small
\begin{aligned}
 \textbf{Input:} \quad  &\mathcal{Q},\mathcal{K},\mathcal{V} \\
 \textbf{Att:} \quad  &\mathcal{A}_i=\text{Softmax}( \frac{\mathcal{Q}\mathcal{W}_i^Q(\mathcal{K}\mathcal{W}_i^K)^T}{\sqrt{d}} ) \\
 \textbf{Head}:  \quad  &\mathcal{H}_i=\mathcal{A}_i\mathcal{V}\mathcal{W}_i^V,\\
 \textbf{Multi-Head:} \quad & \mathcal{H}= [\mathcal{H}_1,\mathcal{H}_2,...,\mathcal{H}_h]\mathcal{W}^H, \\
 \textbf{Output:} \quad  &\mathcal{Z}=\text{LN}(\mathcal{H}+\mathcal{Q}), \\
\end{aligned}
\end{equation}
where $\mathcal{W}_i^Q, \mathcal{W}_i^K, \mathcal{W}_i^V$, $\mathcal{W}_i^H$ are all trainable matrices; $h$ is the head number and $d_h=d/h$; $\mathcal{A}_i$ is the $i$-th attention matrix corresponding to the $i$-th head matrix; $[\cdot]$ is the concatenation operation; and LN is the Layer Normalization. 

\noindent\textbf{Word-to-Patch (W2P) Attention.}
To calculate the W2P alignment, we apply the conventional W2P attention~\cite{ALPRO}:
\begin{equation} \label{equ:W2P}
 \bm{Z}_{W2P}=\textbf{MHA}(\mathcal{Q}=\bm{V},\mathcal{K}=\mathcal{V}=\bm{X}), \\
\end{equation}
where $\bm{V} \in \mathbb{R}^{N_V \times D}, \bm{X} \in \mathbb{R}^{N_X \times D}$ are respectively video and word embedding sets got from two single-modal encoders. By setting the query $\mathcal{Q}$ to the video patch embeddings, Eq.~\eqref{equ:W2P} learns to assign suitable words to each video patch and thus captures the W2P alignment.

\begin{figure}[t]
  \centering
   \includegraphics[width=1\linewidth,trim = 5mm 0mm 5mm 0mm,clip]{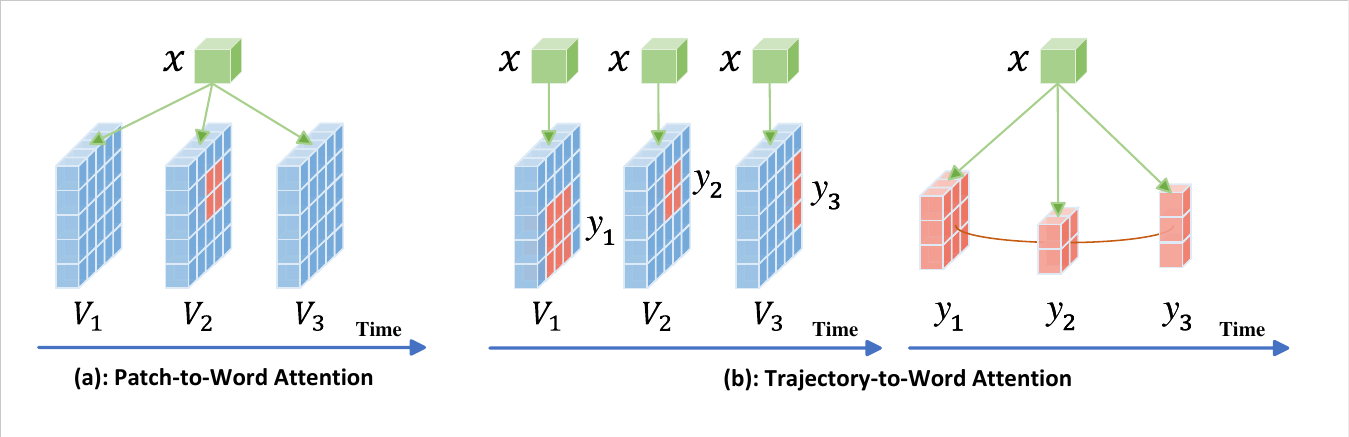}
   \caption{The comparisons between Patch-to-Word (P2W) and Trajectory-to-Word (T2W) attentions, where the green block denotes a query word and the blue blocks denote the video patches. In P2W attention, the word attends over all the video patches and may only concentrate on one frame, while in T2W attention, the salient red parts of each frame are found to construct a trajectory, which are connected by the red lines, \eg, $\bm{y}_1,\bm{y}_2,\bm{y}_3$.}
   \label{fig:T2W}
   \vspace{-0.2in}
\end{figure}

\noindent\textbf{Trajectory-to-Word (T2W) Attention.}
As shown in Figure.~\ref{fig:T2W} (b), we propose the T2W attention that uses two steps to learn the T2W alignment: it first constructs a trajectory for a given word and then uses the word as the query to attend over the trajectory for capturing the associations. For convenience, we introduce how the T2W attention calculates the fusion embedding $\bm{z}_{T2W}$ for a single word $\bm{x}$ and it is straightforward to extend it to a sequence of the words. 

In the first step, T2W attention uses $\bm{x}$ as the query to find the salient parts for each frame and then sequence these parts to construct the trajectory. Assuming $\bm{V}_t$ is the embedding set of the $t$-th frame, the salient part $\bm{y}_{t}$ is got as:
\begin{equation} \label{equ:trajectory}
 \bm{y}_{t}=\textbf{MHA}(\mathcal{Q}=\bm{x},\mathcal{K}=\mathcal{V}=\bm{V}_t). \\
\end{equation}
Then the salient parts at different time frames construct a continuous flow $\bm{Y}=\{\bm{y}_1,...,\bm{y}_T\}$, which is the trajectory of the given word. 

In the second step, to get the trajectory-to-word fusion embedding $\bm{z}_{T2W}$, we treat $\bm{x}$ as the query again while using the trajectory $\bm{Y}$ as the key and value in MHA:
\begin{equation} \label{equ:T2W}
 \bm{z}_{T2W}=\textbf{MHA}(\mathcal{Q}=\bm{x},\mathcal{K}=\mathcal{V}=\bm{Y}). \\
\end{equation}
By Eq.~\eqref{equ:trajectory}, T2W attention finds the salient parts for the given word at each frame, which enforces Eq.~\eqref{equ:T2W} to attend over the \textbf{continuous} frames instead of concentrating the attention only on one or some \textbf{episodic} frames as the previous P2W attentions. In this way, the whole T2W block exploits more temporal contexts to build vision-language associations, which facilitates the video reasoning tasks that usually require the correct recognition of the temporal patterns. Figure.~\ref{fig:T2W} compares P2W and T2W attentions.
\begin{figure*}[t]
  \centering
   \includegraphics[width=1\linewidth,trim = 4mm 3mm 5mm 0mm,clip]{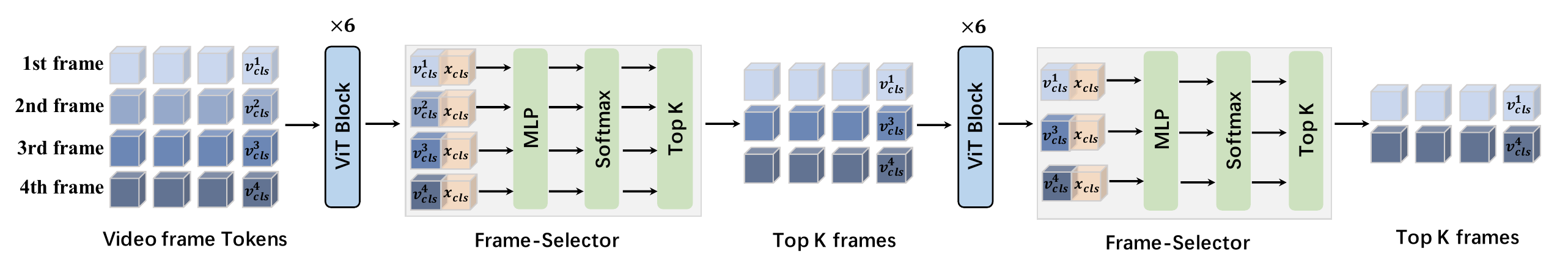}
   \caption{The architecture of the proposed Hierarchical Frame-Selector. At beginning, four frames are input into the video encoder and the first frame-selector behind the 6-th layer filters out the 2-nd frame. At last, only the tokens of the 1-st and 4-th frames are preserved.}
   \label{fig:frame-selector}
   \vspace{-0.2in}
\end{figure*}

\subsection{Hierarchical Frame-Selector}
\label{subsec:hierarchical_frame_selector}
During fine-tuning, to improve the performance, previous studies tend to sample more frames than pre-training for introducing more temporal information. However, on the one hand, using too many frames will dramatically increase the computational resources, which causes infeasible training. On the other hand, in a video, different trajectories have varying graininess, \eg, some trajectories may cover longer frames and some cover shorter. In this way, uniform sampling may introduce redundant temporal information for these longer trajectories and deficient information for the shorter ones. Interestingly, the time duration of the trajectory is also reflected in the corresponding text. In this way, we can use the text to choose suitable frames and then only input them into the cross-modal attention for learning trajectory-word alignments.

Motivated by this, we propose a novel \textbf{Hierarchical Frame-Selector} (\textbf{HFS}) that can gradually select the most relevant video frames to reduce the computational burdens while providing more temporal knowledge. Figure~\ref{fig:frame-selector} illustrates the architecture of our proposed frame-selector module. HFS contains a few frame selector layers which have the same structure with different parameters. For each frame-selector layer, given a sequence of video frame [CLS] tokens $\bm{V}_{cls}=\{\bm{v}^1_{cls},\bm{v}^2_{cls},...,\bm{v}^T_{cls}\} \in \mathbb{R}^{T \times D}$, we concatenate the text [CLS] embedding $\bm{x}_{cls}$ with each $\bm{v}^t_{cls}$: $\bm{f}_{t}=[\bm{v}^t_{cls}, \bm{x}_{cls}] \in \mathbb{R}^{2D} (1 \leq t \leq T)$ and feed them into a scorer network including a FC layer and a softmax layer:
\begin{equation} \label{equ:cls_attn}
 \bm{s} = Softmax(FC(\bm{f})) \in \mathbb{R}^{T \times 1}. \\
\end{equation}
Here, $T$ denotes the number of frames input into the video encoder, and $\bm{s} \in \mathbb{R}^{T}$ is the score vector of T frames. Then for this frame selector, we keep the tokens of $k$ frames with the top-$k$ $s_t$ for the further embedding in the subsequent video Transformer layers. During training, we employ the perturbed maximum method~\cite{pertub} to construct a differentiable Top-K operator. In the video encoder, we totally insert 2 selector layers and insert them in the 6-th and 12-th layers as in Figure~\ref{fig:frame-selector}.

\subsection{Training Objectives}
\label{sec:training_objectives}
To train TW-BERT, as the grey blocks shown in Figure~\ref{fig:framework}, we totally use three losses which are masked language modeling (MLM), video-text matching (VTM), and video-text contrastive loss (VTC).

\noindent\textbf{Masked language Modeling (MLM)~\cite{HERO, ALPRO, VIOLET, all-in-one}.}
MLM aims to predict the masked word tokens given both the video and the text contexts. To get it, we first randomly replace the input text tokens with the [MASK] token with a probability of 15$\%$ and then use the [MASK] embedding output from the cross-modal encoder to predict the masked word by calculating a cross-entropy loss:
\begin{equation} \label{equ:MLM}
\small
\setlength{\abovedisplayskip}{2pt}
\setlength{\belowdisplayskip}{2pt}
    L_{MLM} = \mathbb{E}_{(V,\hat{X}) \sim D}{\rm H}(y^{msk},p^{msk}(V,\hat{X}))
\end{equation}
where $\hat{X}$ is the masked text, {\rm H} is the cross-entropy loss, $y^{msk}$/$p^{msk}$ are the ground-truth/predicted masked tokens.

\noindent\textbf{Video-text Matching (VTM)~\cite{ALPRO, VIOLET, all-in-one}.}
VTM calculates whether the given video and text are matched or not. To get it, for a given video-text pair, we first randomly replace the text with the ones from a different video in the same batch. Then we concatenate the video and text [CLS] embeddings output from the cross-modal encoder and input the concatenated embedding into a binary classifier to judge whether the given video-text pair is matched or not:
\begin{equation} \label{equ:VTM}
\small
\setlength{\abovedisplayskip}{2pt}
\setlength{\belowdisplayskip}{2pt}
    L_{VTM} = \mathbb{E}_{(V,X) \sim D}{\rm H}(y^{vtm},p^{vtm}(V,X))
\end{equation}
where $y^{vtm}$/$p^{vtm}(V,X)$ are the ground-truth/predicted values indexing whether $V$ and $X$ are matched or not.

\noindent\textbf{Video-text Contrastive (VTC)~\cite{ALPRO, all-in-one, objectawareVP, BridgeFormer, MILES, Frozen}.}
As detailed in Section~\ref{sec:approach}, VTC contrasts the outputs of two single-modal encoders to pull close their embedding space to help the subsequent cross-modal encoder build more robust vision-language associations. Suppose $s_{ij}$ denotes the similarity score of the $i$-th video and the $j$-th text, for each video and text, we calculate the softmax-normalized video-to-text and video-to-image similarity as:
\begin{equation}\label{equ:VTC_score}
\small
    p^{v2t}_{ij}(V) = \frac{exp(s_{ij}/\tau)}{\sum_{j}exp(s_{ij}/\tau)}, 
    p^{t2v}_{ij}(X) = \frac{exp(s_{ij}/\tau)}{\sum_{i}exp(s_{ij}/\tau)}
\end{equation}
where $\tau$ is a learnable temperature parameter. Let $y^{v2t}(V)$ and $y^{t2v}(X)$ denote the ground-truth one-hot similarity. This loss contains two symmetric parts, where the left term forces the $i$-th text embedding to be close to the $i$-th video embedding compared with the other texts and the right term has a similar effect:
\begin{equation} \label{equ:VTC}
\small
\setlength{\abovedisplayskip}{1pt}
\setlength{\belowdisplayskip}{1pt}
\begin{aligned}
    &L_{VTC} = \\
    &\frac{1}{2}\mathbb{E}_{(V,X) \sim D}[{\rm H}(y^{v2t}(V),p^{v2t}(V)) + {\rm H}(y^{t2v}(X),p^{t2v}(X))]
\end{aligned}
\end{equation}
In the implementation, we follow~\cite{Align} to use the momentum queue as a continuously-evolving teacher to provide more negative samples.  

\section{Experiments}
\label{sec:experiments}

\subsection{Pre-training Dataset}
Following recent work~\cite{ALPRO, BridgeFormer,MILES,objectawareVP,Frozen}, we pre-train TW-BERT on Google Conceptual Captions (CC3M)~\cite{CC3M} containing 3.3M image-text pairs and  WebVid-2M~\cite{Frozen} containing 2.5M video-text pairs. For CC3M, the image is
treated as a one-frame video data during pre-training.
Note that due to the limited storage and computation resources, we do not use some much larger datasets like HowTo100M~\cite{HowTo100M} containing 136M video-text pairs as~\cite{HERO,VideoCLIP}. Also, we do not distill knowledge from CLIP~\cite{CLIP}, which is pre-trained on 400M image-text pairs, as~\cite{CLIP4Clip}.

\subsection{Downstream Tasks}
\noindent\textbf{Text-to-Video Retrieval.}
($\romannumeral1$) \textbf{MSRVTT} contains 10K YouTube videos with 200K descriptions. We follow~\cite{MSRVTT} to use 9K train+val videos for training and report results on the 1K test split.
($\romannumeral2$) \textbf{DiDeMo}~\cite{DiDeMo} contains 10K Flickr videos annotated with 40K sentences.
($\romannumeral3$) \textbf{LSMDC} consists of 118,081 video clips sourced from 202 movies, where the validation set and the test set contain 7,408 and 1,000 videos.
($\romannumeral4$) \textbf{ActivityNet Caption} contains 20K YouTube videos annotated with 100K sentences. The training set contains 10K videos, and we use val1 set with 4.9K videos to report results.
For MSRVTT and LSMDC, we perform standard text-to-video retrieval. For DiDeMo and ActivityNet Caption, we concatenate all the text captions in the same video as a single query and evaluate paragraph-to-video retrieval. For two tasks, we measure the performances by average recall at K(R@K) and Median Rank on zero-shot and fine-tune setups.

\noindent\textbf{Video Question Answering.}
($\romannumeral1$) \textbf{MSRVTT-QA}~\cite{QA} is built upon videos and captions from MSRVTT~\cite{MSRVTT}, which contains 10K videos with 243K open-ended questions and 1.5K answer candidates.
($\romannumeral2$) \textbf{MSVD}~\cite{MSVD} contains 50K question-answer pairs with 2423 answer candidates. We use standard train/val/test splits for the two tasks, and report accuracy.

\begin{figure*}[t]
  \centering
   \includegraphics[width=1\linewidth]{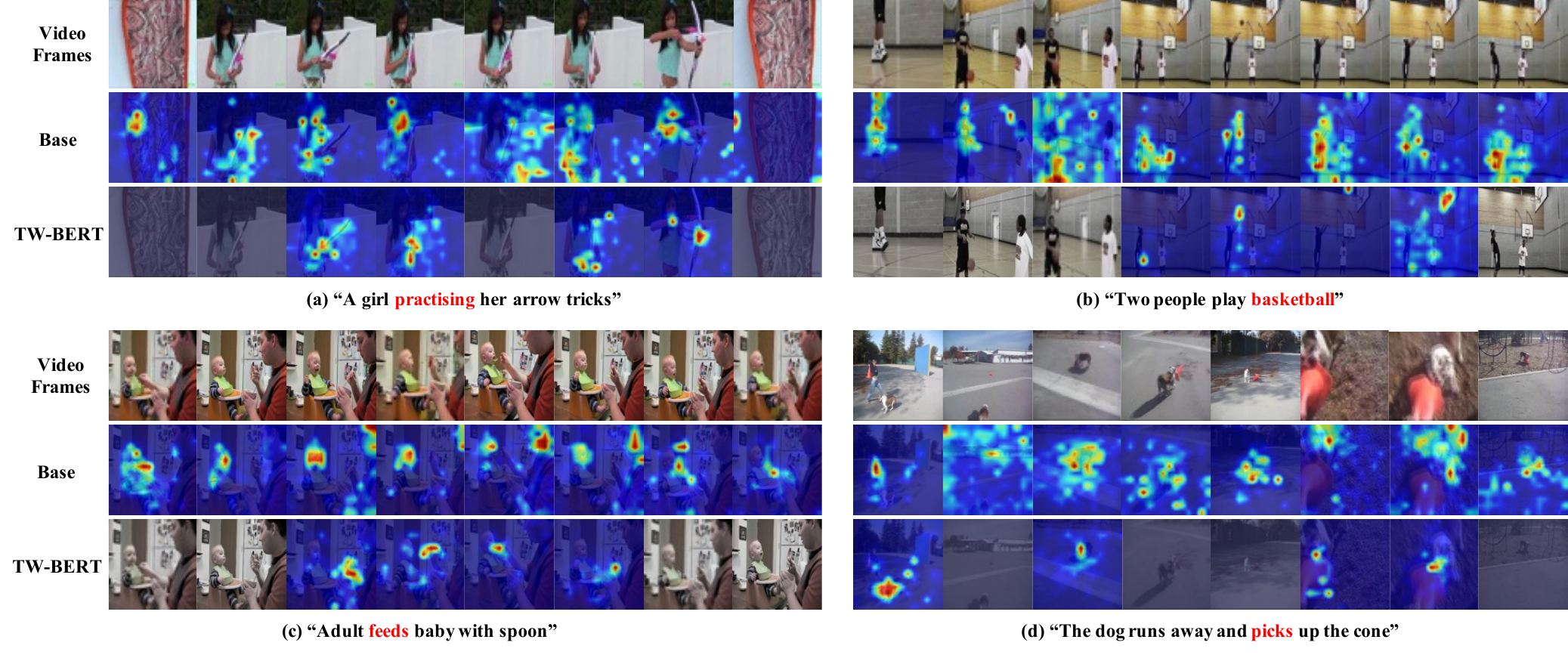}
   \caption{Visualizations of the attention maps from cross-modal encoder. Sample (a) and (b) are from MSRVTT~\cite{MSRVTT}, (c) and (d) are from DiDeMo~\cite{DiDeMo} retrieval dataset. \textbf{TW-BERT} attends to the patches related to given query word by Trajectory-to-Word attention.}
   \label{fig:visualization}
   \vspace{-0.1in}
\end{figure*}

\begin{table*}[!ht]
  \centering
  \caption{Performances of various baselines. R@K and MedR respectively denote recall (\%) with K retrieval efforts and median ranking for retrieved videos where higher R@K and lower MedR indicate better performance.}
  \scalebox{0.95}{
  \begin{tabular}{lccccccccc}
    \toprule[1.5pt]
    \multirow{2}*{Method} &\multicolumn{4}{c}{MSRVTT-ZS} &\multicolumn{4}{c}{DiDeMo-ZS} & MSVD-QA\\ 
    \cmidrule(lr){2-5}  \cmidrule(lr){6-9} \cmidrule(lr){10-10}
     & R@1$\uparrow$ & R@5$\uparrow$ & R@10$\uparrow$ & MedR$\downarrow$ & R@1$\uparrow$ & R@5$\uparrow$ & R@10$\uparrow$ & MedR$\downarrow$ & Acc.\\ 
    \hline
    \textbf{Base} & 25.1 & 46.4 & 57.3 & 7.0 & 26.6 & 52.8 & 62.7 & 5.0 & 47.4\\
    \hline
    \textbf{T2W} & \textbf{26.8} & \textbf{50.0} & \textbf{59.9} & \textbf{5.0} & \textbf{28.4} & 52.9 & \textbf{64.5} & \textbf{4.0} & \textbf{48.2}\\
    \hline
    \textbf{MeanP} & 25.8 & 47.9 & 58.0 & 6.0 & 27.4 & \textbf{53.1} & 64.0 & 5.0 & 48.0\\
    \bottomrule[1.5pt]
  \end{tabular}
  }
  \label{tab:ablation}
  \vspace{-0.1in}
\end{table*}

\begin{table}[!ht]
  \centering
  \caption{Performances comparison with different layer settings of the \textbf{HFS} in fine-tuning.}
  \scalebox{0.9}{
  \begin{tabular}{lccccc}
    \toprule[1.5pt]
    \multirow{2}*{layers} &\multicolumn{4}{c}{MSRVTT-FT} & MSVD-QA\\ 
    \cmidrule(lr){2-5}  \cmidrule(lr){6-6} 
     & R@1$\uparrow$ & R@5$\uparrow$ & R@10$\uparrow$ & MedR$\downarrow$ & Acc.\\ 
    \hline
    \textbf{[2,4]} & 36.7 & 63.2 & 74.9 & 3.0 & 46.2\\
    \hline
    \textbf{[3,6]} & 37.4 & 63.7 & 75.4 & 3.0 & 46.9\\
    \hline
    \textbf{[4,8]} & 38.1 & 64.6 & 76.2 & 3.0 & 47.8\\
    \hline
    \textbf{[5,10]} & 38.2 & 64.9 & 76.3 & 3.0 & 48.3\\
    \hline
    \textbf{[6,12]} & \textbf{38.4} & \textbf{65.1} & \textbf{76.6} & \textbf{3.0} & \textbf{48.5}\\
    \bottomrule[1.5pt]
  \end{tabular}
  }
  \label{tab:HFS2}
  \vspace{-0.1in}
\end{table}

\begin{table}[!ht]
  \centering
  \caption{Performances comparison with different frame settings of the \textbf{HFS} in fine-tuning.}
  \scalebox{0.68}{
  \begin{tabular}{lcccclc}
    \toprule[1.5pt]
    \multirow{2}*{Frame} &\multicolumn{4}{c}{MSRVTT-FT} & \multirow{2}*{Frame} & MSVD-QA\\ 
    \cmidrule(lr){2-5}  \cmidrule(lr){7-7} 
     & R@1$\uparrow$ & R@5$\uparrow$ & R@10$\uparrow$ & MedR$\downarrow$ & & Acc.\\ 
    \hline
    \textbf{F@1} & 35.8 & 63.3 & 74.2 & 3.0 & \textbf{F@8} & 47.2\\
    \hline
    \textbf{F@4-2-1} & 35.9 & 63.2 & 73.9 & 3.0 & \textbf{F@20-14-8} & 47.2\\
    \hline
    \textbf{F@4} & 37.1 & 63.9 & 75.3 & 3.0 & \textbf{F@12} & 47.7\\
    \hline
    \textbf{F@8-6-4} & 37.5 & 64.2 & 75.5 & 3.0 & \textbf{F@24-18-12} & 47.9\\
    \hline
    \textbf{F@8} & 38.1 & 64.9 & 76.0 & 3.0 & \textbf{F@16} & 48.2\\
    \hline
    \textbf{F@20-14-8} & \textbf{38.4} & \textbf{65.1} & \textbf{76.6} & \textbf{3.0} & \textbf{F@32-24-16} & \textbf{48.5}\\
    \bottomrule[1.5pt]
  \end{tabular}
  }
  \label{tab:HFS}
  \vspace{-0.2in}
\end{table}

\subsection{Implementation Details}
We initialize our video encoder by ViT-B/16~\cite{Imagenet} and the text encoder by the first six BERT-Base layers~\cite{bert}. 
For the cross-modal encoder, the self-attentions in all 3 cross-modal attentions (W2P contains 1 and T2W contains 2) are initialized by the last 6 BERT-Base layers~\cite{bert}. The model is trained end-to-end during both pre-training and fine-tuning.

In pre-training, the feature dimension is set to 256 when calculating the contrastive losses and the temperature is set to 0.05. For the momentum queue, the momentum value is 0.995 and the size of the queue is 65,536. The above implementation details follow the recent work~\cite{Align, ALPRO} for a fair comparison. We pre-train the model on CC3M and WebVid-2M for 10 epochs on 8 NVIDIA A100 GPUs where the batch size is 128. We use AdamW~\cite{Adam} optimizer with a weight decay of 0.001 and betas (0.9, 0.98). The learning rate is first warmed-up to 1e-4 and then decays following a linear decay schedule. 

During fine-tuning text-to-video retrieval, We sample 20 frames and use \textbf{HFS} to preserve 8 frames to feed into the cross-modal encoder. The model is trained with both VTC and VTM losses, and we obtain similarity scores from the output of the VTM head during inference. For video question answering, we sample 32 frames per video and preserve 16 frames. Since MSRVTT-QA and MSVD-QA~\cite{QA} are open-ended VQA, in which the answers are in free-form natural language, it is common to convert the task to a classification task by predicting the answer's label. We input the concatenation of the video and question [CLS] tokens into a two-layer MLP~\cite{bert} for calculating the cross-entropy loss. All the fine-tuning experiments are conducted on 8 NVIDIA V100 GPUs.

\subsection{Ablation Studies}
We conduct comprehensive ablation studies to evaluate the effectiveness of the proposed trajectory-to-word (T2W) attention and Hierarchical Frame-Selector (HFS) module. 

\noindent\textbf{Comparing Methods.}
\textbf{Base}: We use a symmetric cross-modal encoder that contains patch-to-word (P2W) and word-to-patch (W2P) attentions. \textbf{T2W}: We replace P2W attention in Base by our T2W attention.  \textbf{MeanP}: Another simple way to consider temporal knowledge is to mean pool the embeddings along the temporal axis. Specifically, we simply average $\{\bm{y}_1,...,\bm{y}_T\}$. These three ablations are implemented using the zero-shot setting. 

\textbf{[L1,L2]}: We insert frame-selector layer into the L1-th and L2-th layers. \textbf{F@T}: We sample T frames into the model. \textbf{F@T1-T2-T3}: We use Hierarchical Frame Selector where ``F@T1-T2-T3'' denotes that T1 frames are originally input into the model and then T2 and T3 frames are respectively remained by the first and second frame-selector layers. All these ablations are implemented using the fine-tuning setting.

\noindent\textbf{Quantitative Results.}
Table~\ref{tab:ablation} compares the performances of diverse baselines. From this table, we can see that T2W outperforms Base, \eg, T2W achieves 3.6\% R@5 improvements on MSRVTT for zero-shot evaluation, which suggests that our T2W attention can exploit more temporal contexts to better solve video-language tasks. 
Also, T2W achieves higher scores than MeanP on different tasks, \eg, T2W achieves 28.4\% of R@1 score in DiDeMo zero-shot text-to-video retrieval task while MeanP only has 27.4\%. Applying mean-pooling strategy indeed fuse all the temporal knowledge, however, simply averaging all $\bm{y}_t$ also means that the contexts \underline{\textit{over the whole time axis}} are used, while lots of objects may only \underline{\textit{span a few frames}} and using all the temporal contexts may introduce trivial or even harmful noises. This proves that our T2W attention can capture more important cross-modal associations in this asymmetric cross-modal encoder case.

We further analyze the impact of the layer settings and frame number on the Hierarchical Frame-Selector module. 
As shown in Table~\ref{tab:HFS2}, frame-selector in deep layers (\eg 6-th and 12-th layers) brings better performances on both retrieval and QA tasks. Moving frame-selector module into shallower layers deteriorates the scores. For example, the accuracy drops 0.9\% when the frame-selector module is placed before the third layer. We hypothesize that discarding frames too early would result in a lack of temporal knowledge in subsequent ViT layers. We thus choose to insert the frame-selector module in the 6-th and 12-th layers in the following experiments.
In terms of frame number, as shown in Table~\ref{tab:HFS}, on the one hand, as the number of frame we sample increases, both retrieval and QA performance obtain improvements. For example, in MSRVTT fine-tuning text-to-video retrieval task, F@20-14-8 outperforms F@8-6-4 by 0.9\% on R@1 or on MSVD video question answering task, F@32-24-16 outperforms F@24-18-12 by 0.6\%, which suggests that more temporal knowledge benefit our model whether using uniform sampling or our proposed \textbf{HFS}. On the other hand, using HFS gets higher scores than uniform sampling, \eg, F@20-14-8 achieves 38.4\% of R@1 score in MSRVTT retrieval task while F@8 only has 38.1\%, or on MSVD video question answering task, F@32-24-16 outperforms F@16 by 0.3\%, which validates the power of HFS module.

\noindent\textbf{Qualitative Results.}
We show the heat maps of the attentions of \textbf{Base} and \textbf{TW-BERT} in Figure~\ref{fig:visualization}. We see that TW-BERT can select the most relevant frames according to the whole text, then it implicitly form a trajectory among these frames for a given query word to avoid over-exploiting the trivial spatial contexts as in Base. For example, in (a), TW-BERT first discards the 1-st and 8-th frames which is completely irrelevant to the text, then tracks the hand of the girl in each frame that it keeps according to the query word ``practising'' while Base cannot avoid impact of the irrelevant frames, and in those relevant frames, it only attends to the larger while trivial regions about the whole body of the girl. Moreover, in (d), TW-BERT keeps the 6-th and 7-th frames for these two frames show the dog's action more clearly. Then TW-BERT attends to the part where the dog is in contact with the cone according to the query word ``picks'' while Base only focuses on the body of the dog.

\begin{table*}[!ht]
  \centering
  \caption{Experiments of text-to-video retrieval on MSRVTT, DiDeMo and LSMDC datasets. ``$\#$PT Pairs'' lists the number of video-text pairs for pre-training. We show results with zero-shot evaluation (top) and fine-tuning evaluation (bottom). R@A is the average of R@1, R@5 and R@10.}
  \scalebox{0.66}{
  \begin{tabular}{llccccccccccccccc}
    \toprule[1.5pt]
    \multirow{2}*{Method} & \multirow{2}*{$\#$PT Pairs} &\multicolumn{5}{c}{DiDeMo} &\multicolumn{5}{c}{LSMDC} &\multicolumn{5}{c}{MSRVTT}\\ 
    \cmidrule(lr){3-7}  \cmidrule(lr){8-12} \cmidrule(lr){13-17}
     & & R@1$\uparrow$ & R@5$\uparrow$ & R@10$\uparrow$ & R@A$\uparrow$ & MedR$\downarrow$ & R@1$\uparrow$ & R@5$\uparrow$ & R@10$\uparrow$ & R@A$\uparrow$ & MedR$\downarrow$ & R@1$\uparrow$ & R@5$\uparrow$ & R@10$\uparrow$ & R@A$\uparrow$ & MedR$\downarrow$\\ 
    \midrule
    Frozen~\cite{Frozen}  & 5.5M & 21.1 & 46.0 & 56.2 & 41.1 & 7.0 & 9.3 & 22.0 & 30.1 & 20.5 & 51.0 & 18.7 & 39.5 & 51.6 & 36.6 & 10.0  \\
    VIOLET~\cite{VIOLET} & 186M & 23.5 & 49.8 & 59.8 & 44.4 & - & - & - & - & - & - & 25.9 & 49.5 & 59.7 & 45.0 & -  \\
    OA-Trans~\cite{objectawareVP} & 5.5M & 23.5 & 50.4 & 59.8 & 44.6 & 6.0 & - & - & - & - & - & 23.4 & 47.5 & 55.6 & 42.2 & 8.0  \\
    ALPRO~\cite{ALPRO} & 5.5M & 23.8 & 47.3 & 57.9 & 43.0 & 6.0 & - & - & - & - & - & 24.1 & 44.7 & 55.4 & 41.4 & 8.0  \\
    BridgeFormer~\cite{BridgeFormer} & 5.5M & 25.6 & 50.6 & 61.1 & 45.8 & 5.0 & 12.2 & 25.9 & 32.2 & 23.4 & 42.0 & 26.0 & 46.4 & 56.4 & 42.9 & 7.0  \\
    MILES~\cite{MILES} & 5.5M & 27.2 & 50.3 & 63.6 & 47.0 & 5.0 & 11.1 & 24.7 & 30.6 & 22.1 & 50.7 & 26.1 & 47.2 & 56.9 & 43.4 & 7.0  \\
    \textbf{TW-BERT} & 5.5M & \textbf{28.4} & \textbf{52.9} & \textbf{64.5} & \textbf{48.6} & \textbf{4.0} & \textbf{14.2} & \textbf{30.4} & \textbf{36.0} & \textbf{26.9} & \textbf{28.0} & \textbf{26.8} & \textbf{50.0} & \textbf{59.9} & \textbf{45.6} & \textbf{5.0}  \\
    \midrule
    Frozen~\cite{Frozen} & 5.5M & 31.0 & 59.8 & 72.4 & 54.4 & 3.0 & 15.0 & 30.8 & 39.8 & 28.5 & 20.0 & 31.0 & 59.5 & 70.5 & 53.7 & 3.0  \\
    VIOLET~\cite{VIOLET} & 186M & 32.6 & 62.8 & 74.7 & 56.7 & - & 16.1 & 36.6 & 41.2 & 31.3 & - & 34.5 & 63.0 & 73.4 & 57.0 & -  \\
    ALPRO~\cite{ALPRO} & 5.5M & 35.9 & 67.5 & 78.8 & 60.7 & 3.0 & - & - & - & - & - & 33.9 & 60.7 & 73.2 & 55.9 & 3.0  \\
    OA-Trans~\cite{objectawareVP} & 5.5M & 34.8 & 64.4 & 75.1 & 58.1 & 3.0 & 18.2 & 34.3 & 43.7 & 32.1 & 18.5 & 35.8 & 63.4 & 76.5 & 58.6 & 3.0  \\
    BridgeFormer~\cite{BridgeFormer} & 5.5M & 37.0 & 62.2 & 73.9 & 57.7 & 3.0 & 17.9 & 35.4 & 44.5 & 32.6 & 15.0 & 37.6 & 64.8 & 75.1 & 59.2 & 3.0  \\
    MILES~\cite{MILES} & 5.5M & 36.6 & 63.9 & 74.0 & 58.2 & 3.0 & 17.8 & 35.6 & 44.1 & 32.5 & 15.5 & 37.7 & 63.6 & 73.8 & 58.4 & 3.0  \\
    \textbf{TW-BERT}  & 5.5M & \textbf{41.8} & \textbf{71.1} & \textbf{81.2} & \textbf{64.7} & \textbf{2.0} & \textbf{21.0} & \textbf{38.8} & \textbf{49.2} & \textbf{36.3} & \textbf{11.0} & \textbf{38.4} & \textbf{65.1} & \textbf{76.6} & \textbf{60.0} & \textbf{3.0}  \\
    \bottomrule[1.5pt]
  \end{tabular}
  }
  \label{tab:retrieval}
  \vspace{-0.2in}
\end{table*}
  
\begin{table}[!ht]
  \centering
  \vspace{-0.05in}
  \caption{ActivityNet Caption with fine-tuning setting.}
  \scalebox{0.83}{
  \begin{tabular}{cc|cccc}
    \toprule[1.5pt]
    Method & $\#$PT Pairs & R@1$\uparrow$ & R@5$\uparrow$ & R@10$\uparrow$ & MedR$\downarrow$ \\ 
    \midrule
    Dense~\cite{ActivityNet} & - & 14.0 & 32.0 & - & 34.0 \\
    FSE~\cite{FSE} & - & 18.2 & 44.8 & - & 7.0 \\
    CE~\cite{CE} & - & 18.2 & 47.7 & - & 6.0 \\
    HSE~\cite{FSE} & - & 20.5 & 49.3 & - & - \\
    Clipbert~\cite{clipbert} & 5.6M & 21.3 & 49.0 & 63.5 & 6.0 \\
    \textbf{TW-BERT} & 5.5M & \textbf{31.7} & \textbf{62.3} & \textbf{74.9} & \textbf{3.0}\\
    \bottomrule[1.5pt]
  \end{tabular}
  }
  \label{tab:activitynet}
    \vspace{-0.15in}
\end{table}

\begin{table}[!ht]
  \centering
  \caption{Experiments of video question answering on MSRVTT and MSVD datasets in top-1 accuracy (\%).}
  \scalebox{0.95}{
  \begin{tabular}{ll|cc}
    \toprule[1.5pt]
    Method & $\#$PT Pairs & MSRVTT & MSVD \\ 
    \midrule
    Clipbert~\cite{clipbert} & 5.6M & 37.4 & -\\
    ALPRO~\cite{ALPRO}  & 5.5M & 42.1 & 45.9\\
    SINGULARITY~\cite{singularity}  & 5.5M & 42.7 & 45.9\\
    LGDN~\cite{LGDN}  & 15.2M & 43.1 & - \\
    SSML~\cite{SSML}  & 100M & 35.1 & 35.1 \\
    JustAsk~\cite{JustAsk} & 69M  & 41.5 & 46.3\\
    MERLOT~\cite{Merlot}  & 180M & 43.1 & -\\
    VIOLET~\cite{VIOLET}  & 186M & \textbf{43.9} & 47.9\\
    \midrule
    \textbf{TW-BERT}  & 5.5M & 43.6 & \textbf{48.5}\\
    \bottomrule[1.5pt]
  \end{tabular}
  }
  \label{tab:qa}
    \vspace{-0.3in}
\end{table}

\subsection{Comparisons with SOTA}
We compare \textbf{TW-BERT} with previous methods on two frequently applied tasks: video-text retrieval (VDTR) and video question answering (VDQA). Table~\ref{tab:retrieval} and~\ref{tab:activitynet} report the performances of VDTR on MSRVTT~\cite{MSRVTT}, DiDeMo~\cite{DiDeMo}, LSMDC~\cite{LSMDC}, and ActivityNet Caption~\cite{ActivityNet}, respectively, where the former three datasets contain both zero-shot and fine-tuning setups and the last one only has fine-tuning setup. Table~\ref{tab:qa} reports the VDQA on MSRVTT~\cite{MSRVTT} and MSVD~\cite{MSVD}. Among the compared models, MILES~\cite{MILES}, BridgeFormer~\cite{BridgeFormer}, OA-Trans~\cite{objectawareVP}, Clipbert~\cite{clipbert} and VIOLET~\cite{VIOLET} are SOTA models proposed in recently 1-2 years. Note that VIOLET~\cite{VIOLET} and ALPRO~\cite{ALPRO} distill knowledge from additional large-scale BERTs while we do not. 
Also, we show the number of pre-training video-text pairs in these tables for more clear comparisons.

From these tables, we can find that when the pre-training data is in the same scale, TW-BERT achieves the best performance compared with all the other models on both VDTR and VDQA. For example, on DiDeMo VDTR, TW-BERT outperforms BridgeFormer by 4.8\% on R@1, or on MSVD VDQA, TW-BERT outperforms ALPRO by 2.6\%. Moreover, compared with the models trained on much more data, TW-BERT can still achieve the best performances on various tasks, \eg, on LSMDC VDTR, TW-BERT outperforms VIOLET by 8.0\% on R@10 or on MSRVTT VDQA, TW-BERT outperforms MERLOT~\cite{Merlot} by 0.5\%. 

Note that the videos in LSMDC are longer than MSRVTT and DiDeMo, which means that the videos in this dataset contain more temporal contexts than the other datasets. Then as shown in Table~\ref{tab:retrieval}, the improvements of TW-BERT over other SOTAs are larger than the improvements on the other datasets. For example, compared with BridgeFormer, TW-BERT has an average 3.5\% improvement in the zero-shot setting, while on DiDeMo dataset, the corresponding average improvement over MILES is only 1.6\%. Such comparisons further validate the effectiveness of TW-BERT in exploiting temporal contexts.

Among these SOTAs, only when compared with VIOLET, which uses an additional large-scale model DALL-E~\cite{DALL-E} and 32 more times pre-training data than ours (180M VS. 5.5M), TW-BERT cannot comprehensively surpass VIOLET on all the tasks. For example, on MSRVTT VDQA, VIOLET achieves 0.3\% higher than TW-BERT, while such marginal improvements are got at the cost of much more training resources. Furthermore, TW-BERT still outperforms VIOLET on the other tasks, \eg, on DiDeMo VDTR, TW-BERT outperforms VIOLET by 9.2\% on R@1 or on MSVD VDQA, TW-BERT outperforms VIOLET by 0.6\%. These comparisons confirm the effectiveness of the proposed TW-BERT.

\section{Conclusion}
\label{sec:conclusion}
We propose a novel \textbf{T}rajectory-\textbf{W}ord BERT (\textbf{TW-BERT}) that builds Trajectory-to-Word alignments for solving video-language tasks. In particular, we introduce an asymmetric cross-modal encoder that contains word-to-patch (W2P) and Trajectory-to-Word (T2W) to capture cross-modal associations. 
Moreover, a novel Hierarchical Frame-Selector (HFS) module is proposed in the fine-tuning stage to filter out irrelevant frames to help the later T2W attention learn better trajectory-word alignments. Extensive experiments across diverse tasks confirm the effectiveness of the proposed TW-BERT.


{\small
\bibliographystyle{ieee_fullname}
\bibliography{egbib}
}

\end{document}